\documentclass[letterpaper, 10 pt, conference]{ieeeconf}  % Comment this line out if you need a4paper

\IEEEoverridecommandlockouts                              % This command is only needed if 
\usepackage[pdftex]{graphicx}
\usepackage{subfigure}
\usepackage{algorithmic}
\usepackage{algorithm}
\usepackage{listings}

\newcounter{myenumi}
\newenvironment{myenumerate}{\begin{enumerate} \setcounter{enumi}{\themyenumi}}{ \setcounter{myenumi}{\theenumi}\end{enumerate}}

\title{\LARGE \bf
SPRITE:\\Stewart Platform Robot for Interactive Tabletop Engagement
}

\author{Elaine Schaertl Short$^{1}$, Dale Short$^{2}$, Yifeng Fu$^{1}$and Maja J Matari\'{c}$^{1}$% <-this % stops a space
\thanks{$^{1}$Department of Computer Science, University of Southern California, Los Angeles, California, USA.
        {\tt\small elaine.g.short@usc.edu}, {\tt\small yifengfu@usc.edu}, and {\tt\small mataric@usc.edu}}%
\thanks{$^{2}${\tt\small short.dale@gmail.com}}%
}

\begin{document}

\maketitle
\thispagestyle{empty}
\pagestyle{empty}

%%%%%%%%%%%%%%%%%%%%%%%%%%%%%%%%%%%%%%%%%%%%%%%%%%%%%%%%%%%%%%%%%%%%%%%%%%%%%%%%
\begin{abstract}

%TODO: update abstract
We present the design of the Stewart Platform Robot for Interactive Tabletop Engagement (SPRITE).  This robot is designed for use in socially assistive robotics, a field focusing on non-contact social interaction to help people achieve goals relating to health, wellness, and education. We describe a series of design goals for a tabletop, socially assistive robot, including expressive movement, affective communication, a friendly, nonthreatening, and customizable appearance, and a safe, robust, and easily-repaired mechanical design. 

\end{abstract}

%%%%%%%%%%%%%%%%%%%%%%%%%%%%%%%%%%%%%%%%%%%%%%%%%%%%%%%%%%%%%%%%%%%%%%%%%%%%%%%%
\section{INTRODUCTION}
SPRITE is designed for use in socially assistive robotics (SAR), an area of robotics in which robots help people achieve goals through non-contact social interaction, especially in the areas of health, wellness, and education \cite{Feil-Seifer2005}.  Socially assistive robots have been developed for a variety of uses, including to help patients recover from stroke-related disability \cite{Swift-Spong2015a}, to aid in social skill training for children with autism spectrum disorders \cite{Feil-Seifer2011}, to encourage adolescents to exercise \cite{Swift-Spong2016}, to help young children learn number concepts \cite{Clabaugh2015}, and to coach older adults in seated exercises \cite{Fasola2013}, among others.

Based on design goals for a tabletop socially assistive robot, we developed a platform whose mechanical design and software are publicly available to the scientific community for research applications\footnote{Hardware: \textit{http://robotics.usc.edu/$\sim$sprite}}\footnote{Software: \textit{https://github.com/interaction-lab/cordial-public}}.  The platform primarily uses off-the-shelf and 3D-printed physical components with off-board computation and platform-independent mobile phone faces that allow the computational components to be readily replaced as more powerful hardware becomes available.  The total cost of the mechanical components of the robot is affordable in numbers facilitating conducting large-scale, multi-site research studies.  We describe the results of several pilot studies in which participants were asked about their perceptions of SPRITE.  Participants found the robot non-threatening and attractive, and requested a customizable appearance, suggesting that we were successful in achieving our design goals.

\begin{figure}[h]
\begin{center}
\subfigure[Robot as "Chili" (dragon) \label{fig:chili}]{\includegraphics[height=0.5\linewidth]{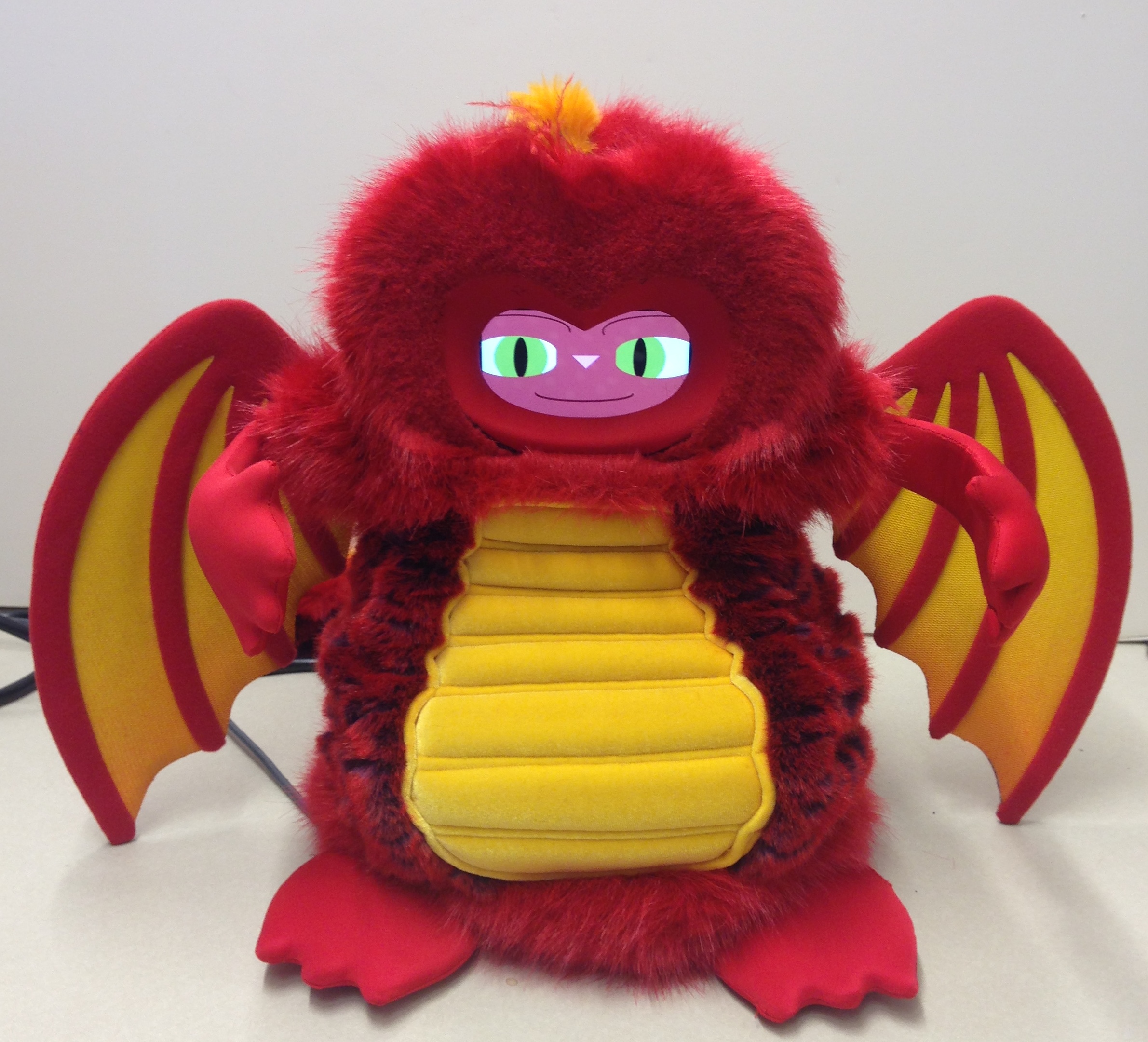}}
~
\subfigure[Robot as "Kiwi" (stylized bird) \label{fig:kiwi}]{\includegraphics[height=0.5\linewidth]{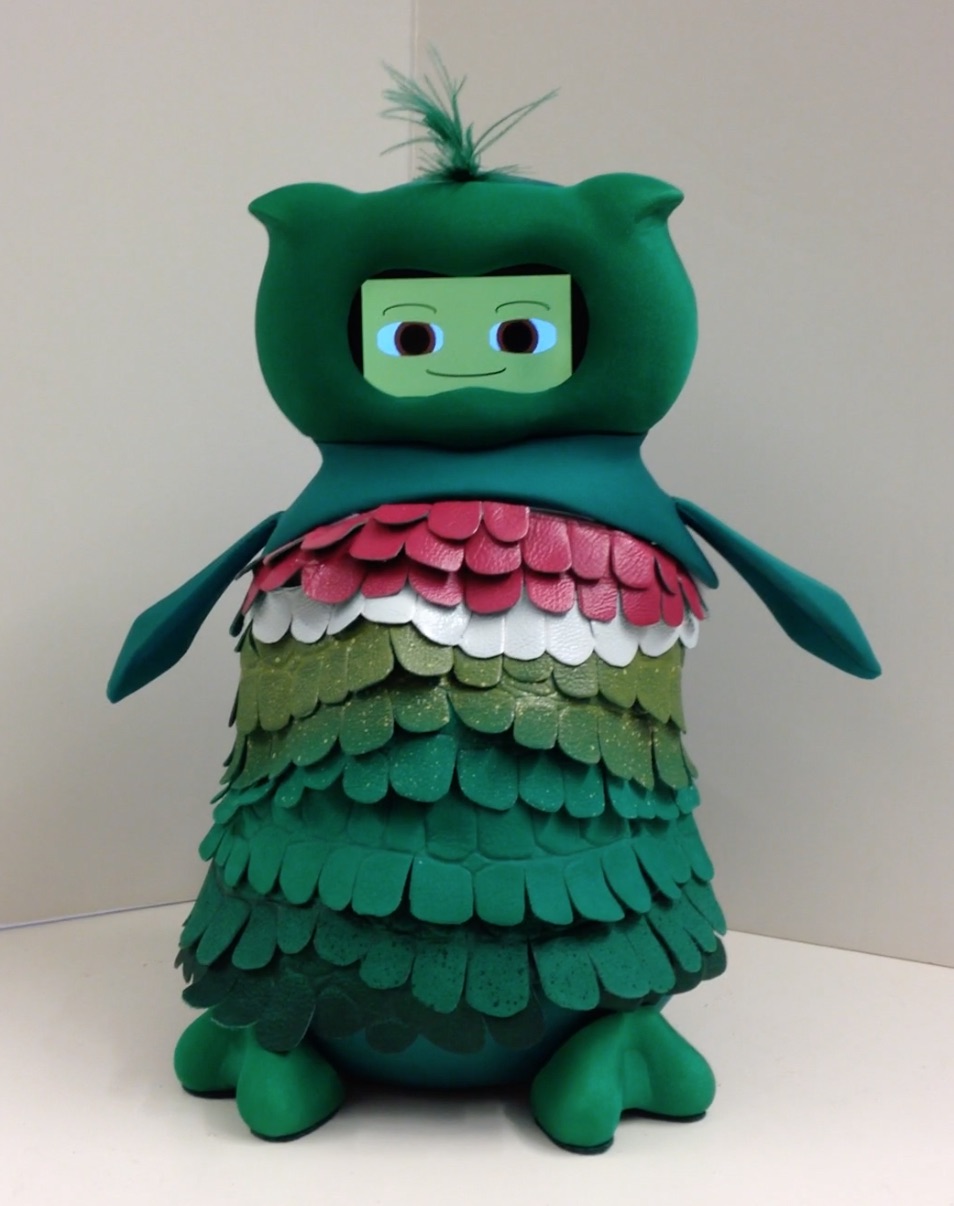}}
~
\subfigure["Chili" face\label{fig:chili_face}]{\includegraphics[height=0.24\linewidth]{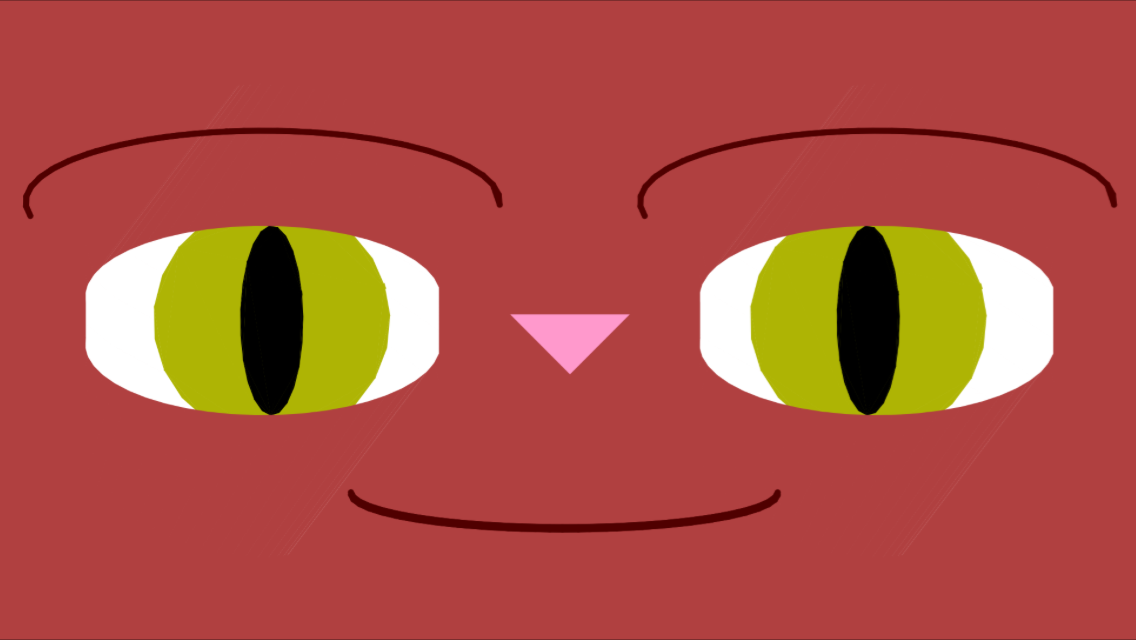}}
~
\subfigure["Kiwi" face\label{fig:kiwi_face}]{\includegraphics[height=0.24\linewidth]{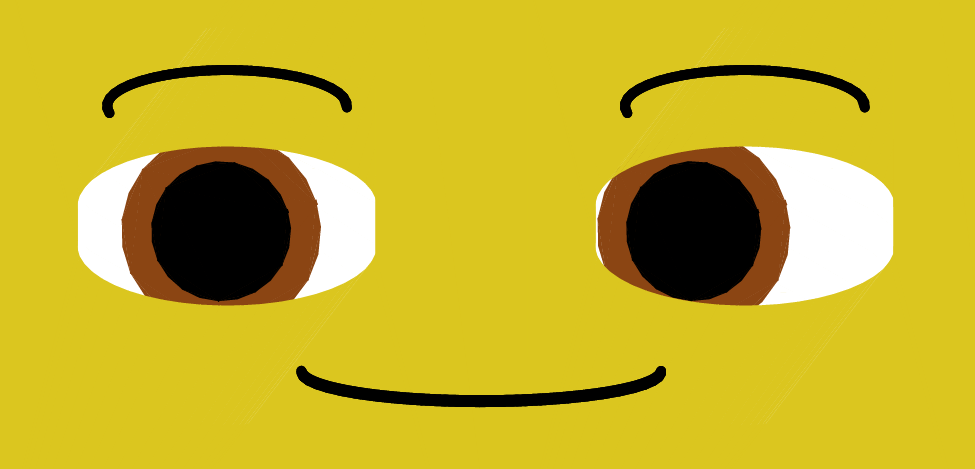}}
\caption{Custom robot skins used in research \label{fig:research_skins}}
\end{center}
\end{figure}

To deploy the robot, we developed several ``skins'', including two professionally-designed, proprietary skins (Figure \ref{fig:research_skins}) and a category of toddler-clothing-based skins that can be shared with the research community (Figure \ref{fig:sprite_clothes}).

In addition to the robot hardware, we present CoRDial, the Co-Robot Dialogue system: a software stack implemented in the Robot Operating System (ROS) \cite{quigley2009ros} that includes controllers for the SPRITE as well as robot-independent nodes for playing synchronized speech and behaviors such as body animations and appropriate mouth movements.  The robot's face, part of the CoRDial software stack, can be displayed on any device with a JavaScript-enabled browser and can be used to provide any robot with an expressive display-based face.

\section{RELATED WORK}
The appearance of the robot, especially the ``Chili'' skin (Figure \ref{fig:chili}), is based in part on the concept for the DragonBot robot \cite{Setapan2012}, a tabletop dragon robot with an inverted-delta-platform mechanical design \cite{Clavel1988}.  Additionally, there have been a number of other tabletop robots for social interaction research such as iCat, used to teach children chess \cite{Leite2008}; Travis, a music-listening companion \cite{Hoffman2012}; Keepon, one of several small robots designed for children with autism \cite{kozima2007cri, Scassellati2012}; Maki, a 3D-printed robot \cite{Hello-robo}; and Jibo, a commercial product still under development as of this publication \cite{jibo}.  Our design occupies a similar niche, but was developed specifically for SAR research with more degrees of freedom in the body than Maki, Keepon, Travis, DragonBot, and Jibo and more degrees of freedom in the animated face than Keepon, iCat, and Maki.  SPRITE is designed with the needs of the human-robot interaction (HRI) research community in mind, with offboard computation to enable the latest computing platforms to be used to control the robot and a robust and easy-to-repair physical design.

\section{DESIGN GOALS \label{sec:reqs}}
Based on our USC Interaction Lab’s SAR research involving various target user populations and contexts, we developed a set of goals for the robot's design.  Because they use social interaction to motivate, coach, and teach, socially assistive robots need to be able to express a range of emotional responses, such as being happy when a patient in physical rehabilitation finishes their exercises or sad when a child has trouble with a math problem.  In order to engage in appropriate social behavior, the robots should also be able to engage in basic nonverbal behavior, such as nodding or using gaze to direct the user's attention. To summarize:

\begin{myenumerate}
\item The robot must be capable of expressive movement. \label{reqs:movement}
\item The robot must be capable of affective communication. \label{reqs:affective}
\end{myenumerate}

A socially assistive robot used in research might be used to study interactions with a wide variety of users, from children with developmental challenges to older adults with motor impairments.  For these users, the robot's basic size and shape should be non-threatening and appealing so as to motivate interaction.  A robot that appeals to children might not appeal to older adults, so the ability to customize the robot's appearance to the user population is important.  Finally, SAR applications often involve long-term interactions with users outside of the direct supervision of an experimenter, so the robot must be as safe as possible.  To summarize:

\begin{myenumerate}
\item The robot should have a friendly, inviting, and non-threatening appearance for both adult and child users.\label{reqs:appearance}
\item The robot's appearance should be customizable for different target users. \label{reqs:customizable}
\item The robot should be safe to interact with, for novice as well as experienced users.\label{reqs:safe}
\end{myenumerate}

Additionally, because SAR involves non-contact interactions, the robot does not need to manipulate objects.

\subsection{Engineering Considerations}
In addition to the requirements listed above, because SPRITE is designed to be a research platform, it needs to function through multiple development cycles, user studies, and demonstrations, and should remain useful for several years, as computational and electronics component technologies improve.  The robot also needs to integrate with existing software; as with the hardware, the software should be modular, allowing individual components to be replaced as better versions become available.  This introduces several additional considerations:
\begin{myenumerate}
\item The robot should be robust and easy to repair or upgrade. \label{reqs:robust}
\item The robot should be sufficiently inexpensive to allow replication in multi-site studies. \label{reqs:inexpensive}
\item The robot should include control software that allows for rapid development of human-robot interactions. \label{reqs:software}
\item The robot's software should integrate with other robotics software, such as the Robot Operating System (ROS) \cite{quigley2009ros}. \label{reqs:ROS}
\end{myenumerate}

\begin{figure}[t]
\begin{center}

\vspace{.2cm}

\includegraphics[width=0.7\linewidth]{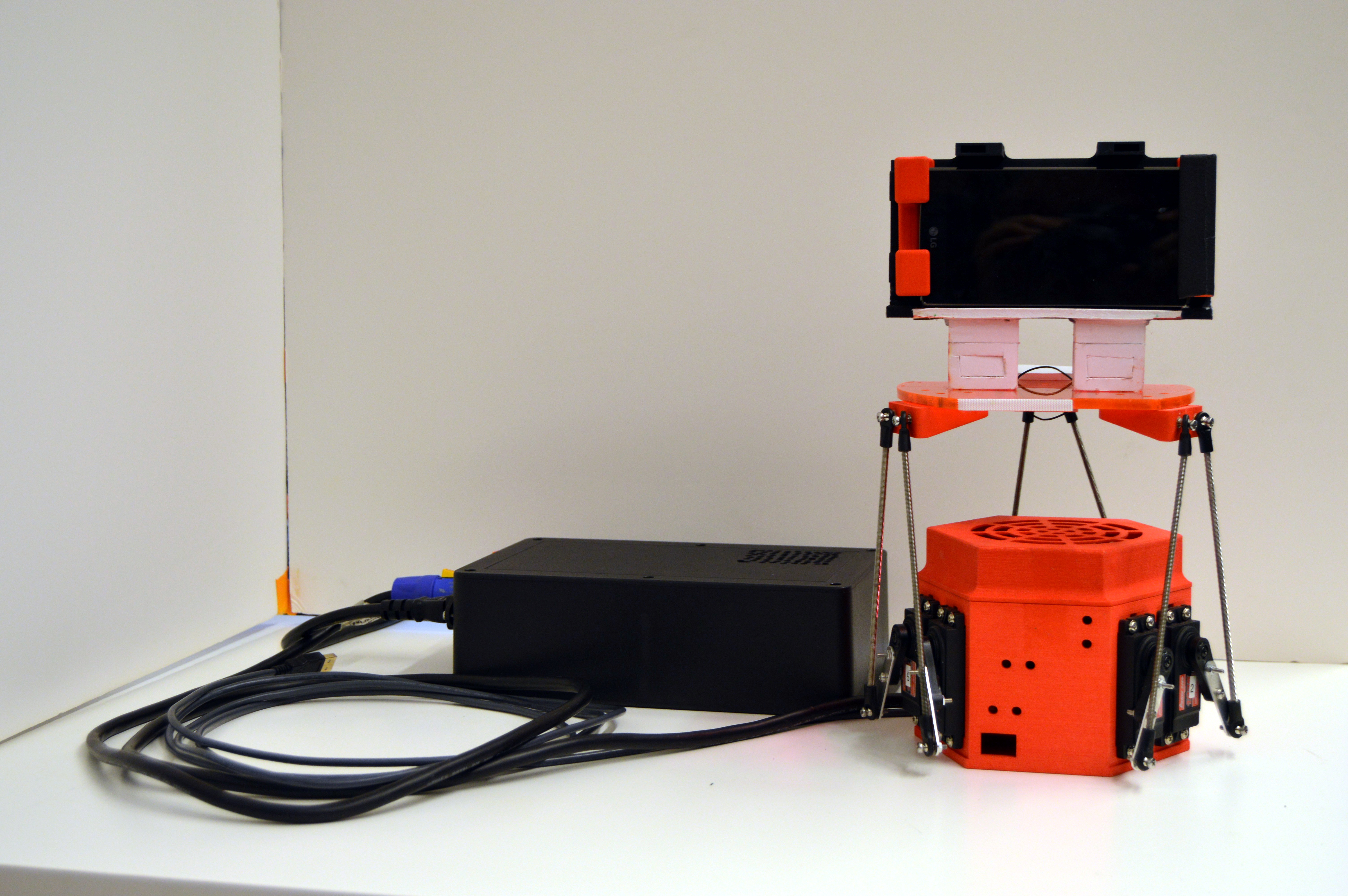}
\caption{Internal hardware of the SPRITE, with added ``neck''\label{fig:hardware}}
\end{center}
\end{figure}

\begin{figure}[t]
\begin{center}
\includegraphics[width=0.7\linewidth]{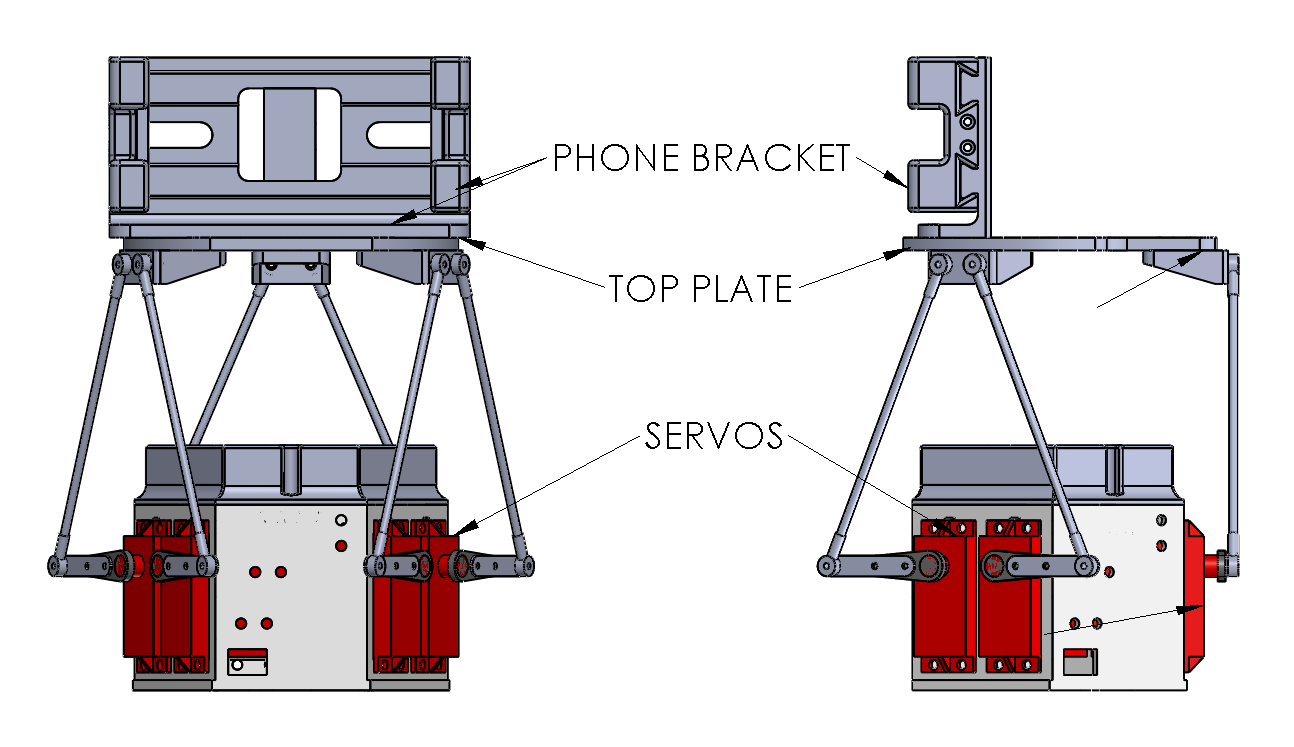}
\caption{Schematic of the internal SPRITE design\label{fig:hardware_drawing}}
\end{center}
\end{figure}

\section{ROBOT DESIGN}
SPRITE's materials and basic design allow for flexibility in the external appearance of the robot, facilitating customization for different target applications.

\subsection{Physical Design}
The underlying design of SPRITE is based on the rotary, six degree-of-freedom Stewart platform \cite{Stewart1965}. On the platform is mounted a bracket that holds a mobile phone that displays the robot’s face.  In interactions, the robot’s appearance can be customized with a skin that covers the mechanical components. The robot is powered by an external power supply.  Instructions for assembling the robot and power supply, including models for 3D printing, are freely available to the research community\footnote{By correspondence with the authors and released publicly since December 2016}.  More details about the design of the robot are found in the following sections.

\begin{figure}[t]
\begin{center}

\vspace{.2cm}

\subfigure[z=+4cm (moved up) \label{fig:z+}]{\includegraphics[height=0.4\linewidth]{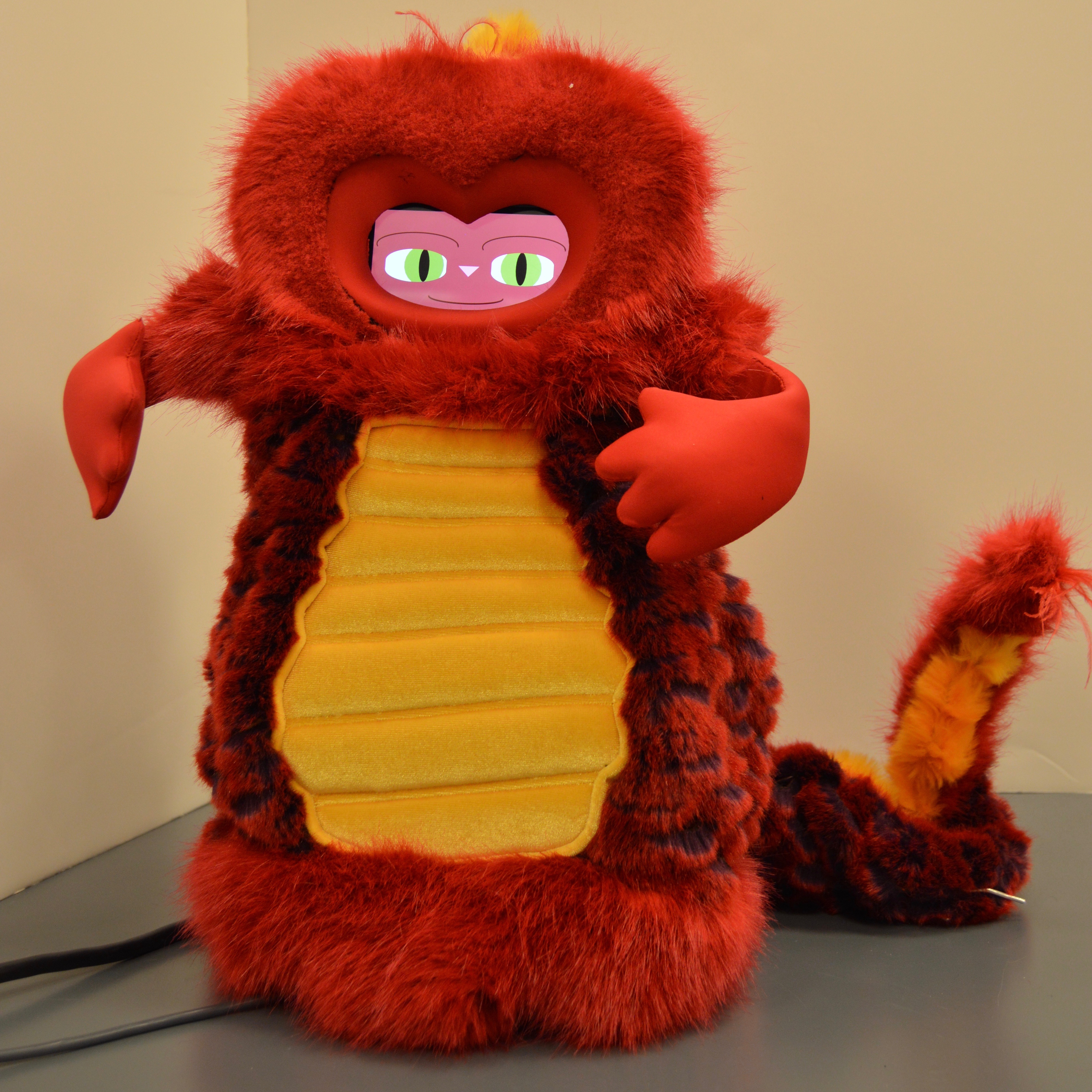}}
~
\subfigure[z=-4cm (moved down) \label{fig:z-}]{\includegraphics[height=0.4\linewidth]{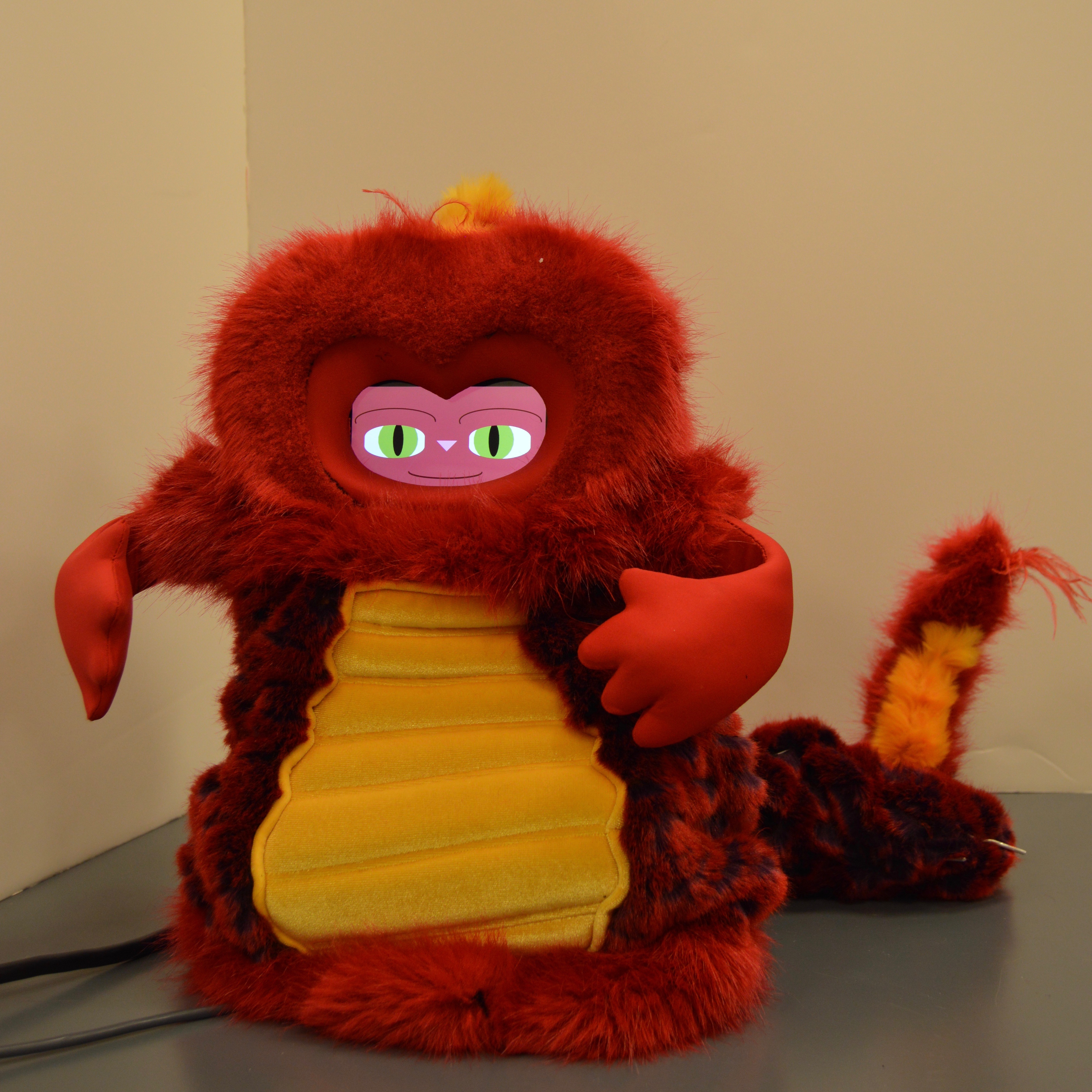}}
~
\subfigure[pitch=+30 degrees \newline (look up)\label{fig:pitch+}]{\includegraphics[height=0.4\linewidth]{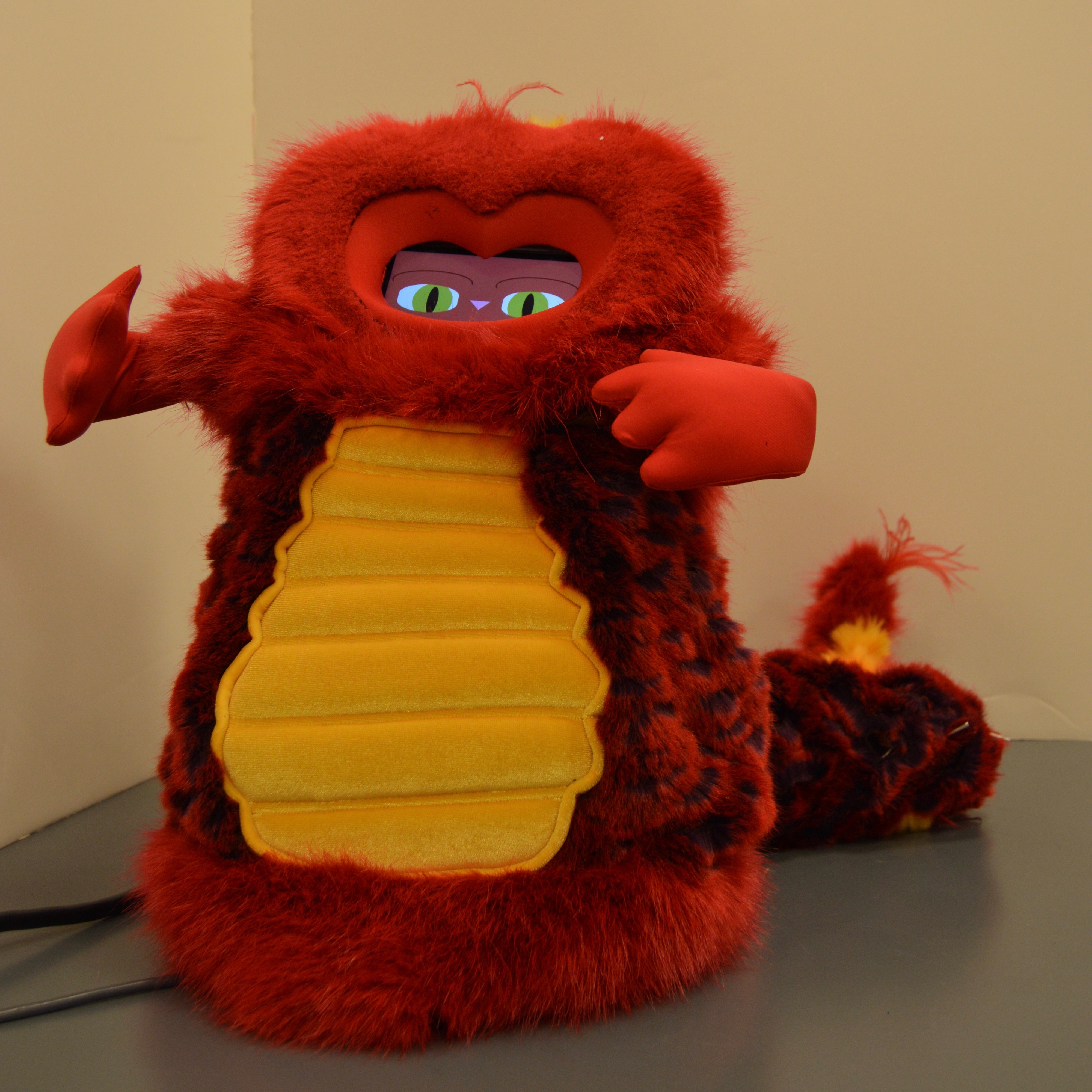}}
~
\subfigure[pitch=-30 degrees \newline (look down) \label{fig:pitch-}]{\includegraphics[height=0.4\linewidth]{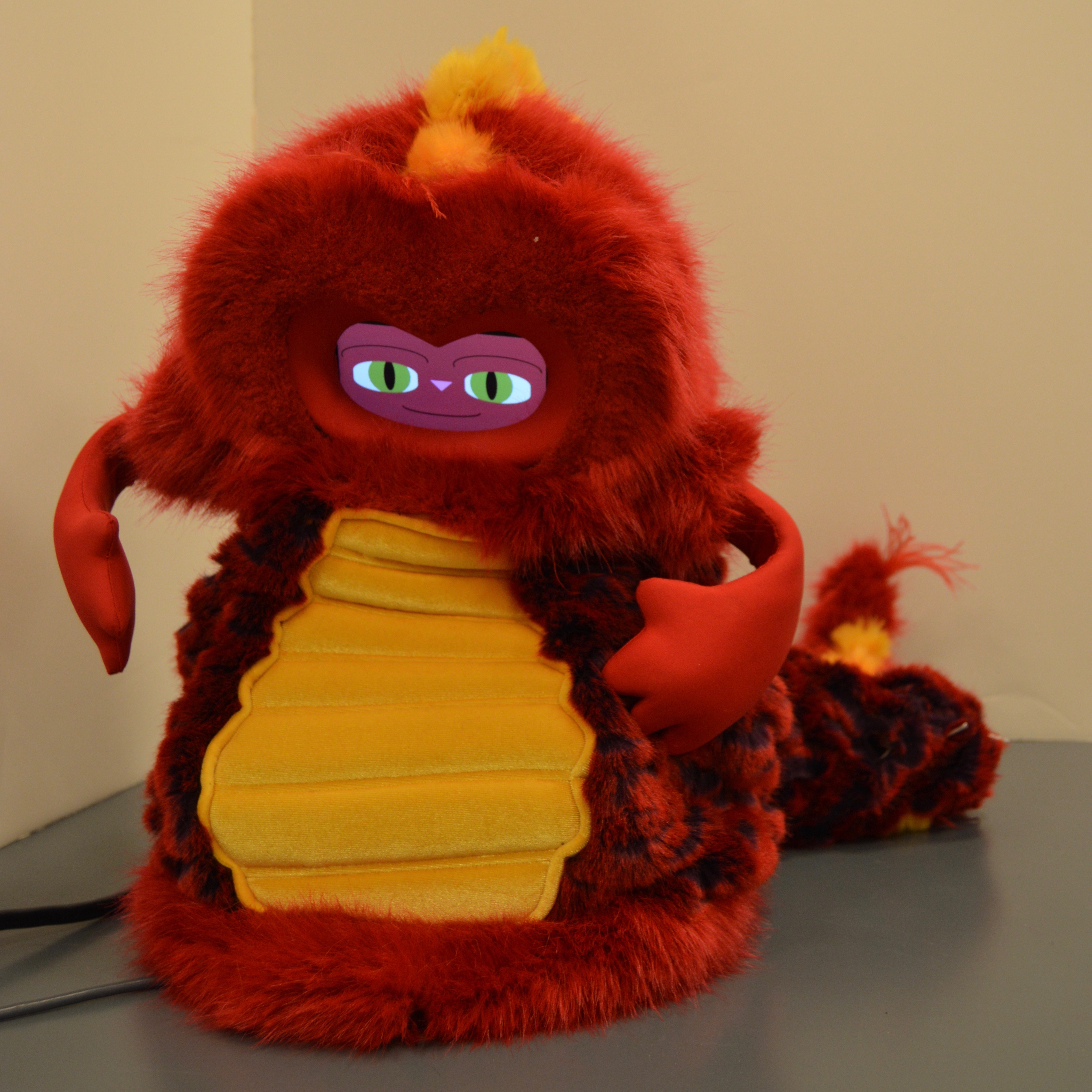}}
~
\subfigure[roll=+30 degrees \newline (tilted head left)\label{fig:roll+}]{\includegraphics[height=0.4\linewidth]{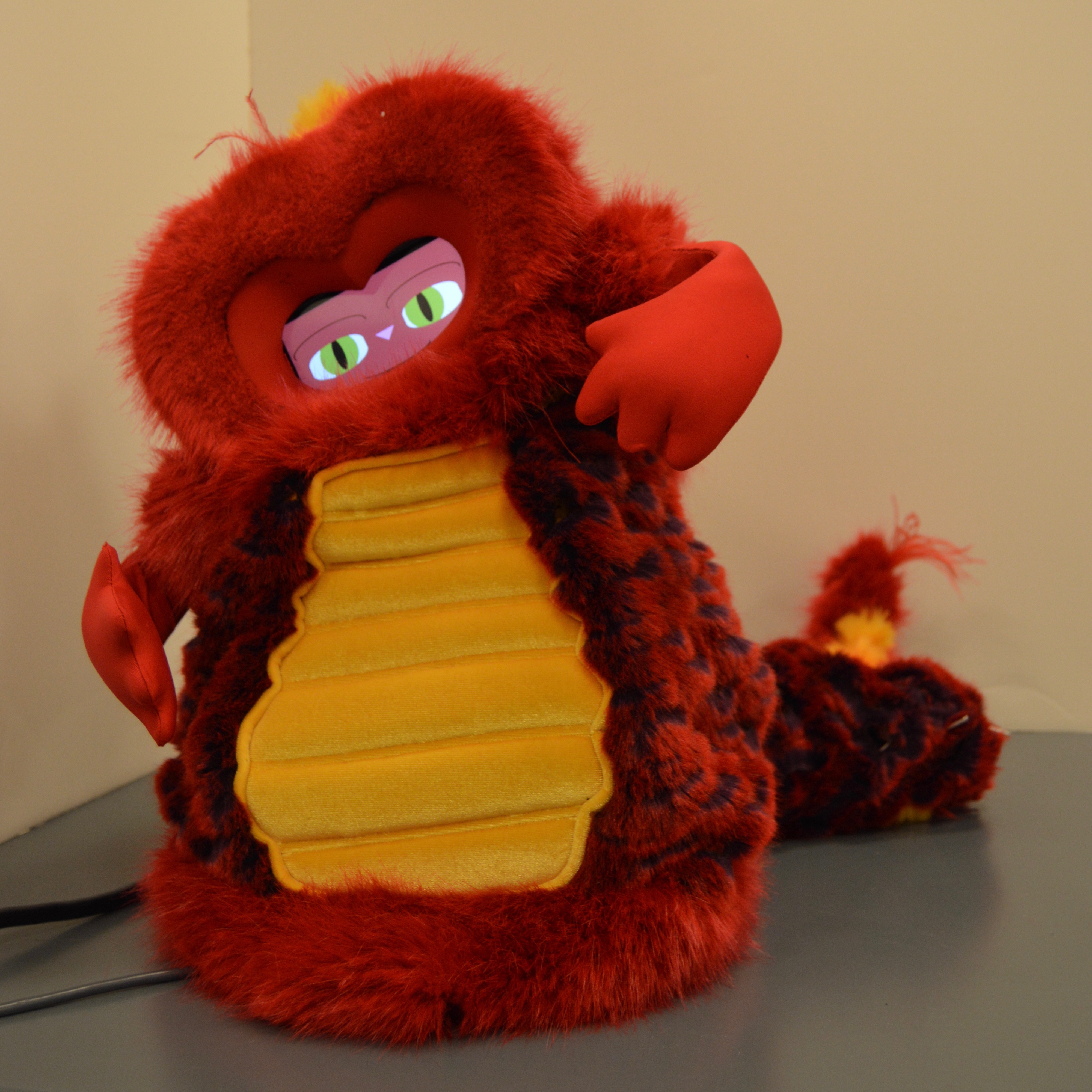}}
~
\subfigure[roll=-30 degrees \newline (tilted head right) \label{fig:roll-}]{\includegraphics[height=0.4\linewidth]{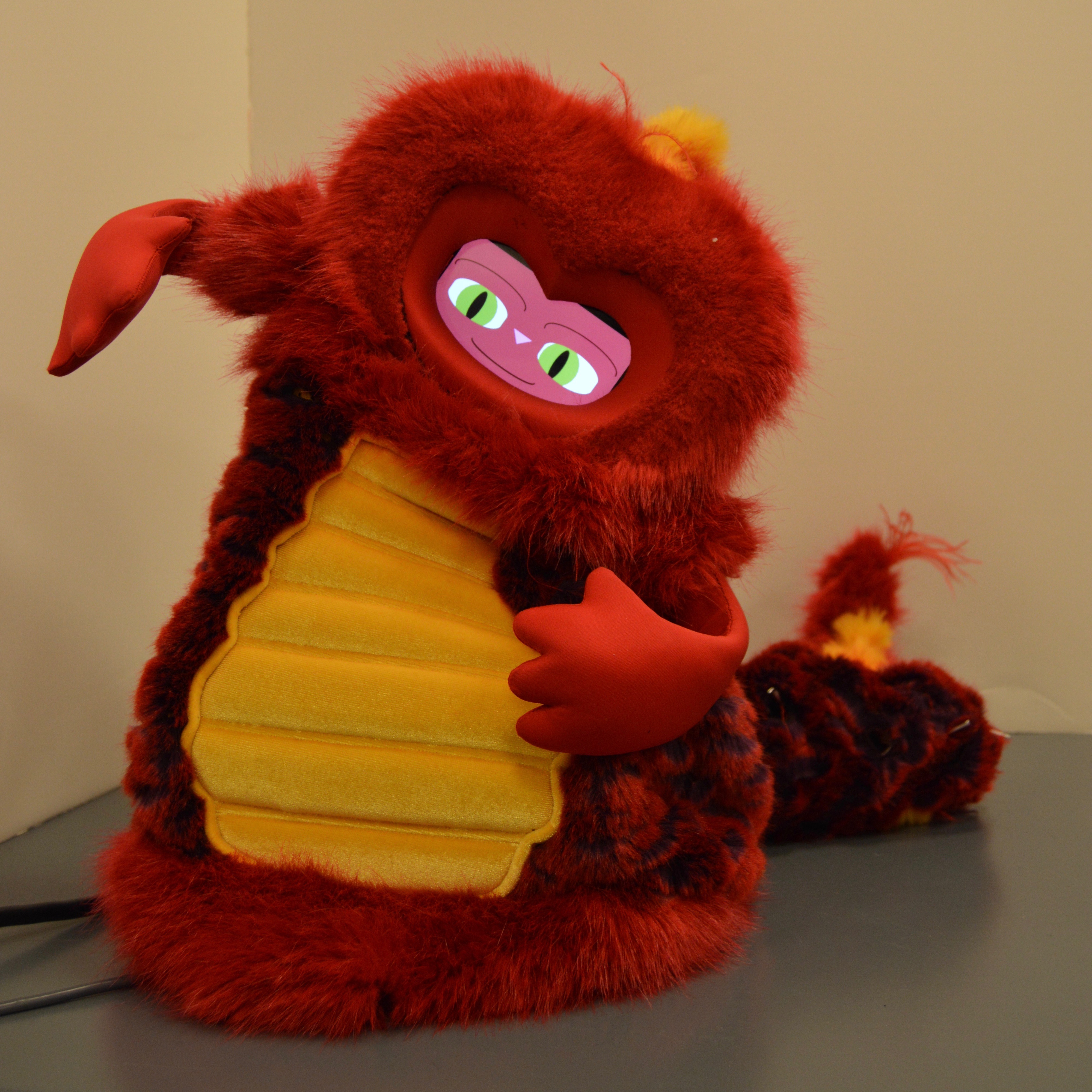}}
\caption{SPRITE robot movement \label{fig:movement}}
\end{center}
\end{figure}

\subsubsection{Dimensions and Workspace}
The Stewart platform design gives the robot's top platform six degrees of freedom: x, y, z, roll, pitch, and yaw. The robot is approximately 30cm tall, fitting easily on a standard desk-sized table. The robot can move approximately 5cm in any direction, and can reach angles of up to around 45 degrees from the vertical axis (see Figure \ref{fig:movement}). The separate power supply is approximately 15cm by 30cm, and with a 1m cord, and can be placed away from the main robot body.

\subsubsection{Hardware} % side-view line drawing of robot goes here
Six metal gear-driven servo motors actuate the top platform and are housed within a 3D-printed hexagonal base.  A laser-cut top plate is attached to an adjustable 3D-printed phone bracket that can accommodate most contemporary phones.  
The locking push-pull connector between the robot and power supply prevents the robot from being accidentally unplugged.  All electrical components are enclosed in the base
away from where a user might access.  The cost of the robot's mechanical parts, with high-torque servos and 3D-printed parts, is under \$1500 at the time of preparation of this manuscript.

\subsubsection{Appearance}
\begin{figure}[t]
\begin{center}

\vspace{.2cm}

\subfigure[]{\includegraphics[height=0.6\linewidth]{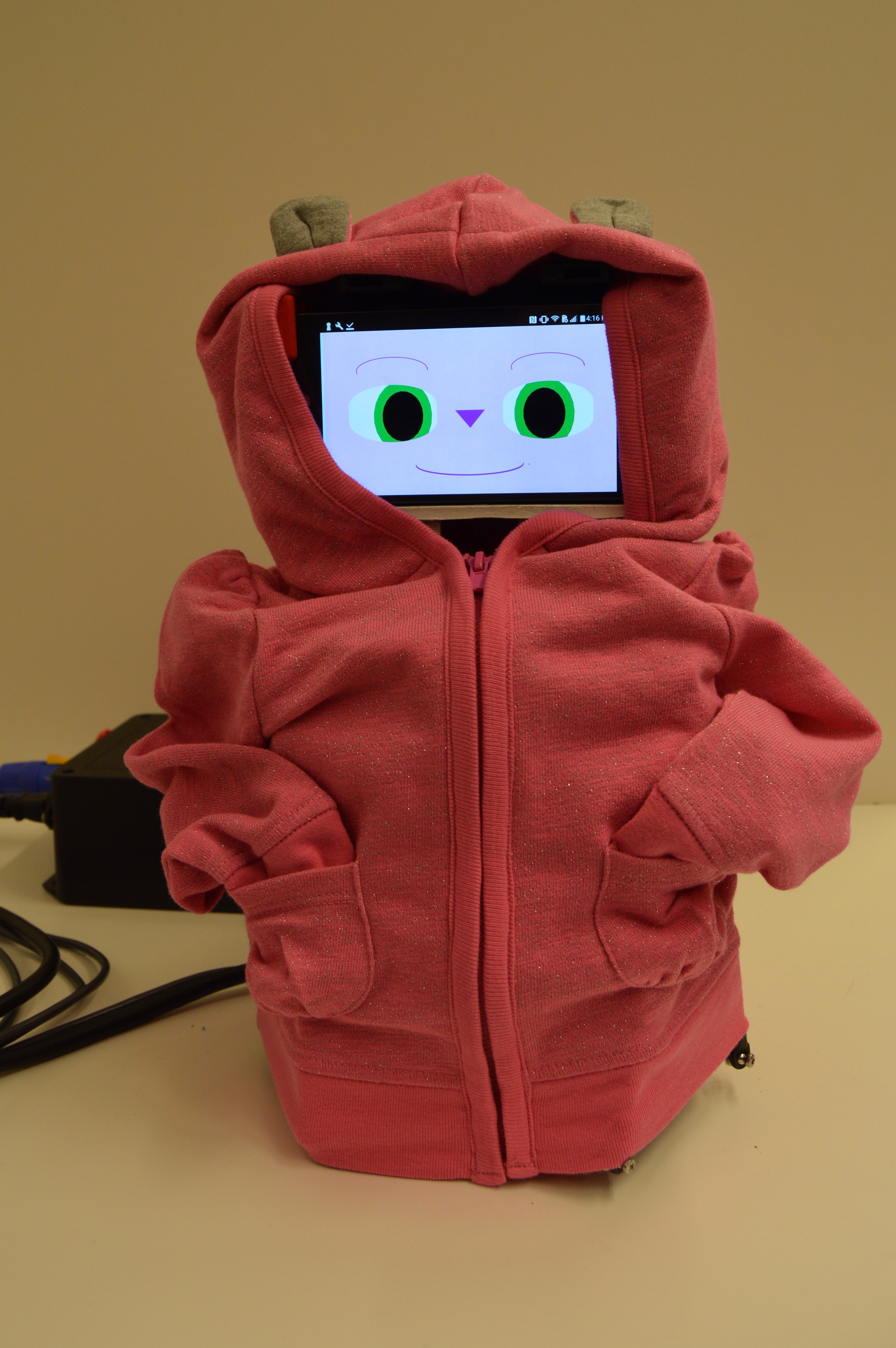}}
~
\subfigure[]{\includegraphics[height=0.6\linewidth]{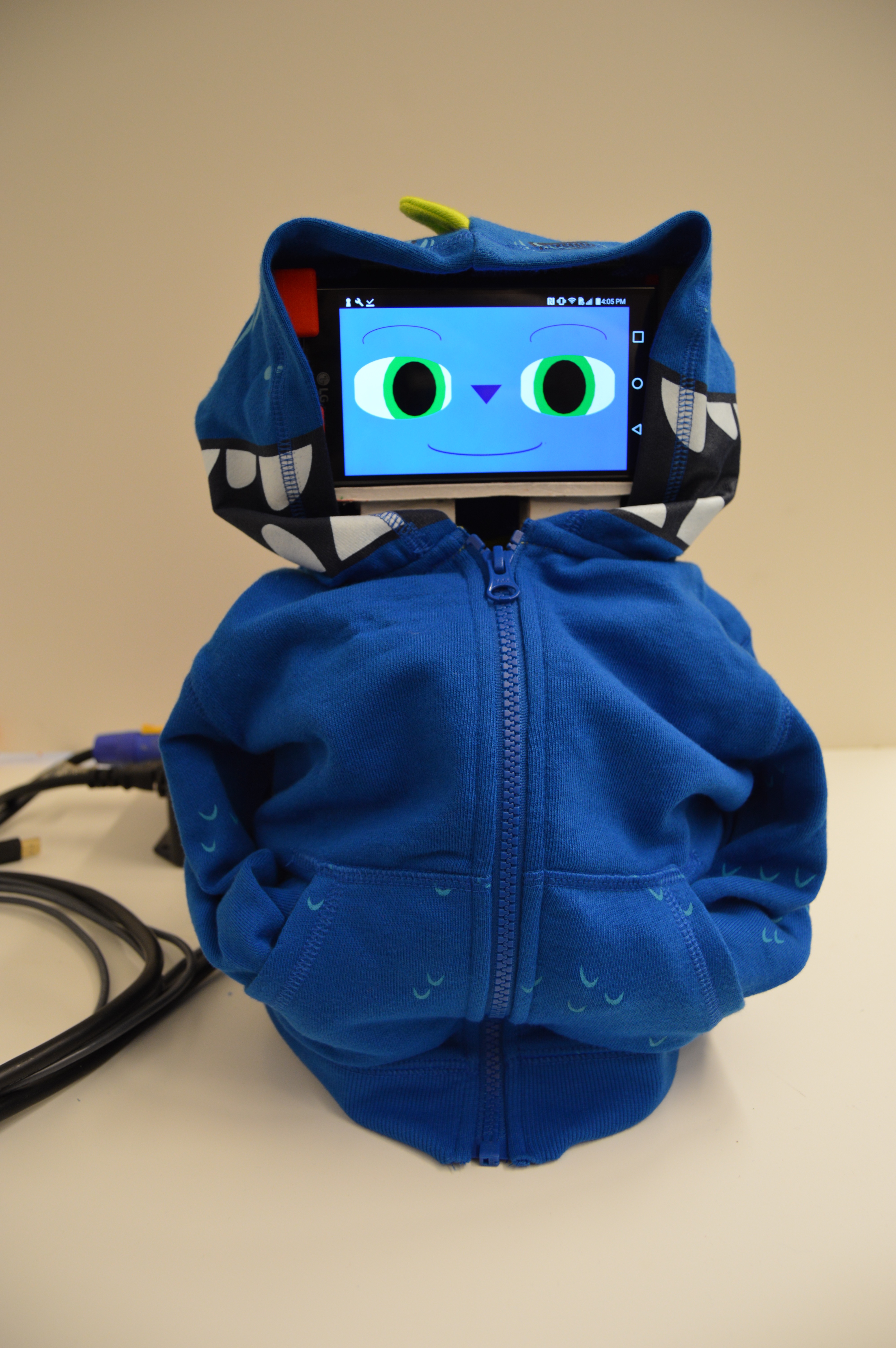}}
\caption{Robot in toddler clothing for customized appearance \label{fig:sprite_clothes}}
\end{center}
\end{figure}

The appearance of the robot is modifiable either through custom-built skins, as seen in Figure \ref{fig:research_skins}, or using toddler-sized (24M/2T, US sizing) hooded sweatshirts, which allow for a wide range of appearances.  Using purposeful choices of face and clothing colors can be used to control the robot's apparent gender, personality, and other characteristics (Figure \ref{fig:sprite_clothes}) depending on the relevance for the target population, or customize the robot to the preferences of an individual user.

\subsection{Software}
The SPRITE robot platform is controlled by CoRDial, the Co-Robot Dialogue system we developed, made freely available to the research community\footnote{By request from the authors; released publicly as of December 2016}. The software stack is designed for the SPRITE robot, but includes a number of robot-independent components as well.  The software components, described in detail in this section, are as follows:
\begin{enumerate}
\item Motor controller (Python) \label{components:motor}
\item ROS interface for keyframe-based animations (Python/ROS) \label{components:animations}
\item Speech with synchronized animations and speech-related mouth movements (visemes) (Python/ROS) \label{components:speech}
\item Browser-based face for mobile phone or computer (JavaScript/ROS) \label{components:face}
\item ROS interface for face and TF tracking (Python/ROS) \label{compontents:face_ros}
\end{enumerate}

Components \ref{components:motor} and \ref{components:animations} are specific to SPRITE, while the remaining components could be used across a wide variety of robots, provided nodes are implemented to translate the robot-independent messages into the correct behavior.

\subsubsection{SPRITE Motor Controller and Keyframe Animations}
The motor control board connects via USB to a computer; all computation is done off-board.  The rotary Stewart platform has a relatively simple inverse kinematic model; motor positions are calculated through the intersection of the circles formed by the servo horns and the push rods.  Software limits are implemented at both the motor control level (by limiting maximum and minimum ticks) and the kinematic control level (by preventing the sending of invalid poses to the motor controller). 

In order to allow researchers to develop animations for the robot, we additionally provide a ROS node that takes a JSON-based keyframe specification and plays the scripted animations.  Interpolation is done using Bezier curves.  The keyframe player node can play facial behaviors as well as robot body movements.

\subsubsection{Synchronized Speech and Movements}
CoRDial's central node is the speech player node, which takes a string that is either the text of a dialogue action (script with tags for behaviors) or an index into a file of saved dialogue actions.  When a request is received, the node plays the corresponding audio, either cached or from a local or remote text-to-speech server, sends the appropriate visemes (mouth positions associated with speech sounds) and expressions to the face, and sends the appropriate movements to the body to synchronize with the speech.  To simplify development of robot interactions, a Python class interface is provided which allows the speech to be triggered with a single line of code.  CoRDial can support multiple robots on a single computer, enabling multi-robot, multi-human interactions.  
 
 \subsubsection{Browser-Based Face}
The face of the robot is implemented in JavaScript, using a 3D animation framework\footnote{three.js \texttt{http://threejs.org}} 
that allows the robot's eyes to be modeled as 3-dimensional spheres.  Communication to the robot's face is done via ROS with the Robot Web Tools \cite{Toris2015}.  In addition to sending a limited set of visemes, the user can activate action units in the face, designed to be analogous to the action units of the Facial Action Coding System (FACS) \cite{ekman1978manual}. The 3D model of the eyes enables the controller to direct the robot's gaze to any point in the three dimensional workspace of the robot, with appropriate vergence by rotating the spheres towards the point.  The 2D projection of this 3D model gives the proper appearance to the eyes on the screen.  If the point is provided as a ROS transform, the ROS node associated with the face will automatically maintain the robot's gaze on the desired point, allowing the eye movement to lead the body movement of the robot while maintaining the gaze target. The colors, pupil shape, and face element sizes (except for the eyes) are fully customizable on the face.  Two examples can be seen in Figure \ref{fig:research_skins}.

\subsubsection{Using CoRDial}
When the CoRDial nodes are running on a computer with access to cloud services (for text-to-speech), it takes only a few lines of code to get the robot to say ``Hello world!'' while doing a keyframe-scripted ``happy dance'' animation with synchronized visemes displayed on the face.  On an offline system, a few additional steps allow the user to download speech audio with synchronized visemes.

This simple interface allows for rapid development of HRI studies, including quickly iterating through wording in robot speech or changing speech online.  Robot speech can also be generated prior to an interaction, allowing for deployment offline or where latency in cloud-based text-to-speech is a problem. 

\section{DISCUSSION}
The design of the SPRITE robot platform addresses a number of goals, as described in Section \ref{sec:reqs}.  These goals, developed from expertise in socially assistive robotics research, address both the capabilities and design of the robot, and the engineering considerations important for research deployments of SAR systems.  The goals and insights are summarized in this section.

\subsection{Design Goals: Expressive Movement and Affective Communication}
SPRITE is designed to be capable of expressive movement (Requirement \ref{reqs:movement}) and affective communication (Requirement \ref{reqs:affective}).  The Stewart platform design and high-torque servos allow the robot to move quickly and smoothly, and keyframe animation with Bezier curve interpolation allows for rapid design of new animations and behaviors.

\subsection{Design Goals: Size and Appearance}
The small size and colorful appearance of the robot makes it non-threatening to both adults and children  (Requirement \ref{reqs:appearance}).  The appearance is customizable (Requirement \ref{reqs:customizable}), allowing researchers to tailor the robot to the specific domain being studied.  Our pilot studies support the importance of personalization; when asked about in-home use of the robot, many participants specifically wanted to be able to customize the robot's appearance.

\subsection{Design Goals: Safety}
A number of features are included to ensure that the system is safe for user interaction (Requirement \ref{reqs:safe}). There is an emergency stop button on the power supply, allowing power to the motors to be quickly cut if necessary. The software is not affected by an emergency stop event since the motor controller board remains powered via USB connection to the computer. The connector between the robot and power supply is designed to have no exposed electrical components and cannot be easily pulled out. 

\subsection{Design Goals: Cost and Performance}
The robot's design uses off-the-shelf and 3D-printed components, making the system easy to repair and upgrade (Requirment \ref{reqs:robust}). The total cost of these components is under \$1500\footnote{At the time of publication.}, making the robot inexpensive enough for replication across multiple sites and multiple in-home deployments (Requirement \ref{reqs:inexpensive}).  

\subsection{Design Goals: Software}
In addition to the hardware, we developed CoRDial, a software stack that includes the SPRITE control software as well as robot-independent components.  At the highest level, the robot can be controlled with a few lines of Python, enabling rapid development and iteration of human-robot interactions (Requirement \ref{reqs:software}). The robot's software is integrated with the Robot Operating System (ROS) in a modular design that enables individual components to be modified or replaced depending on the deployment context and improvements in the state-of-the-art (Requirement \ref{reqs:ROS}).

\section{CONCLUSIONS}
We presented the design of a tabletop socially assistive robot, SPRITE (the Stewart Platform Robot for Interactive Tabletop Engagement), as well as a software stack, CoRDial (the Co-Robot Dialogue system), for controlling the robot, which includes several robot-independent components.  The mechanical design and software stack are freely available to the research community.  

%%%%%%%%%%%%%%%%%%%%%%%%%%%%%%%%%%%%%%%%%%%%%%%%%%%%%%%%%%%%%%%%%%%%%%%%%%%%%%%%

%%%%%%%%%%%%%%%%%%%%%%%%%%%%%%%%%%%%%%%%%%%%%%%%%%%%%%%%%%%%%%%%%%%%%%%%%%%%%%%%

%%%%%%%%%%%%%%%%%%%%%%%%%%%%%%%%%%%%%%%%%%%%%%%%%%%%%%%%%%%%%%%%%%%%%%%%%%%%%%%%
%\section*{APPENDIX}


\section*{ACKNOWLEDGMENT}
The authors thank Clarity Schaertl Short for her assistance with robot design and assembly, and Rhianna Lee, Kara Douville, and Shayna Goldberger for their help with robot behavior programming. This work is supported by the National Science Foundation (Expeditions in Computing IIS-1139148, CNS-0709296, CBET-1548502, REU Supplements, and GRFP).

%%%%%%%%%%%%%%%%%%%%%%%%%%%%%%%%%%%%%%%%%%%%%%%%%%%%%%%%%%%%%%%%%%%%%%%%%%%%%%%%

%References are important to the reader; therefore, each citation must be complete and correct. If at all possible, references should be commonly available publications.

\bibliographystyle{IEEEtran}
\bibliography{icra2017}
%\begin{thebibliography}{1}
%\end{thebibliography}

\end{document}